**Title:**

**The Power of the Pareto Front: Balancing Uncertain Rewards for Adaptive Experimentation in scanning probe microscopy**

**Authors:**

Yu Liu[1*] and Sergei V. Kalinin[1,2*]

[1] Institute for Advanced Materials and Manufacturing, Department of Materials Science and Engineering, University of Tennessee, Knoxville, Tennessee, 37996 USA
[2] Physical Sciences Division, Pacific Northwest National Laboratory, Richland, Washington, 99354, USA

* Corresponding author: yliu206@utk.edu, sergei2@utk.edu

**Abstract:**

Automated experimentation has the potential to revolutionize scientific discovery, but its effectiveness depends on well-defined optimization targets, which are often uncertain or probabilistic in real-world settings. In this work, we demonstrate the application of Multi-Objective Bayesian Optimization (MOBO) to balance multiple, competing rewards in autonomous experimentation. Using scanning probe microscopy (SPM) imaging, one of the most widely used and foundational SPM modes, we show that MOBO can optimize imaging parameters to enhance measurement quality, reproducibility, and efficiency. A key advantage of this approach is the ability to compute and analyze the Pareto front, which not only guides optimization but also provides physical insights into the trade-offs between different objectives. Additionally, MOBO offers a natural framework for human-in-the-loop decision-making, enabling researchers to fine-tune experimental trade-offs based on domain expertise. By standardizing high-quality, reproducible measurements and integrating human input into AI-driven optimization, this work highlights MOBO as a powerful tool for advancing autonomous scientific discovery.

# I. Introduction

Automated scientific discovery is rapidly emerging as a transformative research paradigm, reshaping experimental methodologies through the integration of automated instrumentation, AI-driven decision-making, and multi-tool workflows [1, 2]. This evolution ranges from the automation of individual scientific instruments [3-5] to the full-scale integration of experimental platforms within self-driving laboratories [6-17]. By enabling autonomous hypothesis testing, adaptive experimentation, and real-time optimization, these systems have the potential to significantly accelerate discoveries across various scientific domains [18-21].

A fundamental requirement for active discovery workflows is the definition of optimization targets or reward functions that drive the iterative learning process [18]. These reward functions form the foundation of autonomous workflows, guiding experimental decisions and facilitating interoperability among multiple tools in complex research environments. Designing effective reward functions is particularly critical in applications that involve multi-objective trade-offs, such as materials synthesis [7, 14, 18, 22, 23], high-throughput screening [24], and imaging [5, 25].

However, in many cases, reward functions are inherently uncertain or probabilistic. For instance, in thin-film growth, Raman line intensities may be converted into a scoring metric [26]. In automated experimentation involving scanning probe microscopy (SPM) and scanning transmission electron microscopy, uncertainty is introduced through complex measurement processes [27-32]. In most cases, rather than using a single reward function to define the desired experimental outcome, multiple rewards provide a more natural and intuitive framework for human operators. Typically, each reward function captures a distinct aspect of the desired result, making multi-objective optimization particularly valuable for complex, expensive black-box optimization problems. These problems arise in diverse fields, including scientific experimentation [33], materials discovery [15, 34], and hyperparameter tuning in machine learning [35-37].

Multi-Objective Bayesian Optimization (MOBO) provides an effective approach to optimizing multiple, potentially conflicting objectives while efficiently navigating the parameter space. It leverages Gaussian processes (GP) to model both the mean and uncertainty of expensive objective functions. An acquisition function is then used to balance exploration and exploitation, guiding the selection of subsequent experimental parameters to maximize overall performance.

Rather than identifying a single optimal solution, MOBO seeks to determine a set of trade-off solutions, forming a Pareto front [35, 36, 38]. On Pareto front, no solution can be improved in one objective without incurring a cost in another. When two reward functions have overlapping optimal parameter regions, as illustrated in Figure 1a-b, their Pareto front collapses to a single point in parameter space, resulting in a trivial Pareto solution as shown in Figure 1e. Conversely, if the optimal parameters for two rewards differ, trade-offs become necessary, leading to a non-trivial Pareto front, as depicted in Figure 1f-g. Notably, even if two rewards share an overlapping optimal solution, introducing a third reward with distinct optimal parameters necessitates trade-offs among all three rewards in the joint Pareto front.

SPM is an essential tool for investigating materials and nanoscale phenomena. Among its various imaging and functional spectroscopy modes, tapping mode (TM) is the most widely used [39-44]. However, achieving high-quality SPM images requires extensive tuning of microscope parameters, demanding significant time and expertise [45-47]. Automating the optimization of SPM at an expert level not only ensures standardized and reproducible high-quality imaging but also aligns with the principles of Findability, Accessibility, Interoperability, and Reusability (FAIR), making high-quality SPM data more broadly available to the scientific community.

In this work, we explore the application of multi-objective Bayesian optimization (MOBO) to autonomous experimentation in SPM. We demonstrate that MOBO effectively balances multiple uncertain reward functions, ensuring FAIR (Findable, Accessible, Interoperable, and Reproducible) access to high-quality experimental data. Furthermore, MOBO provides insights into the interdependencies among different reward functions and facilitates human-in-the-loop decision-making by enabling researchers to tailor trade-offs along the Pareto front for specific experimental goals. We illustrate these concepts through the automated optimization of tapping mode in SPM, where three reward functions are derived from either the underlying physics of the system or heuristic criteria. Our results show that MOBO rapidly optimizes control parameters, yielding high-quality and reproducible scans. Additionally, analysis of the Pareto front offers deeper insight into the relationships between competing rewards and provides a framework for integrating human expertise into the decision-making process.

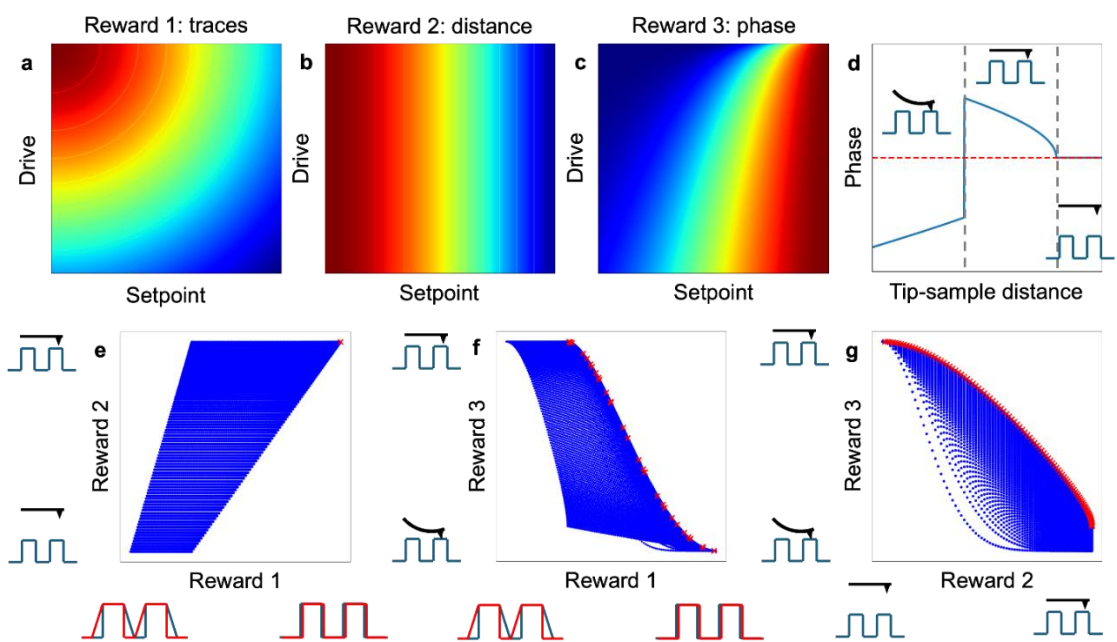

**Figure 1. Rewards design for MOBO of tapping mode in SPM | a,** reward of traces is designed to quantify how well the trace and retrace scan lines are aligned. This reward is expected to favor large drive amplitude and small setpoint parameters. **b,** reward of distance is designed to assess

how close the tip is to the sample surface. This one is computed based on the absolute height of the probe and usually favors small setpoins. **c,** reward of phase is defined to reflect how well the scanning is kept in the attractive mode. It's computed by counting how many phase pixels are in the attractive mode with values above the free-air phase and usually favors small drive amplitudes and large setpoints. Here the expected distribution of rewards is only shown in the drive amplitude vs. setpoint space for easier visualization. **d,** phase as a function of tip-sample distance in theory. When the tip is very far from the sample surface, the phase stays close to the free-air phase (indicated by the red horizontal dashed line). When the probe gets closer and interacts with sample surface in the attractive mode, the phase remains above the free-air phase. When the probe is so close to the sample surface that it's working in repulsive mode, the phase will be below the free-air phase. **e,** expected Pareto scatter plot for reward of trace and reward of distance. The control parameters that give good alignment of trace and retrace are expected to give small tip-sample distance. Therefore, the expected Pareto front collapses into a trivial Pareto point. **f-g,** the reward of phase is competing with the reward of trace and reward of distance in the parameter space. Thus, there are expected trade-offs between them. The cartoons on the axes of e-g show the physical meaning of the extremes of each reward.

## II. Define multiple objectives/rewards for the tapping mode SPM

To define reward functions suitable for optimizing tapping mode across a diverse range of materials, environments, and scales, we focus on identifying common features of high-quality TM scans. First, the trace and retrace of the height channel should be in close agreement and ideally follow the sample's true height profile. However, since the ground-truth sample height profile is generally inaccessible, we define a second key characteristic: the probe should be positioned as close to the sample as possible to minimize the tip-sample distance. Finally, the phase values should consistently remain above the free-air phase (Figure 1d) to ensure scanning occurs in the attractive regime [47-49]. Operating in this regime reduces tip-sample interactions, thereby lowering the risk of tip and sample damage.

$$\text{Reward 1} = -\log \left( \Sigma_i \frac{|h_{trace} - h_{retrace}|}{(h_{trace} + h_{retrace})N} \right) \quad (1)$$

$$\text{Reward 2} = -\log \left( \Sigma_i \frac{N_{\theta < \theta_{free}}}{N} \right) \quad (2)$$

$$\text{Reward 3} = -\log(h_{min} - h_{global\ min}) \quad (3)$$

$$\text{Reward 4} = -\log \left( \Sigma_i \frac{(2 - P(h_{trace}, h_{retrace}))}{N} \right) \quad (4)$$

where $h_{trace}$ and $h_{retrace}$ are the trace and retrace height scan lines, $N$ is the total number of pixels in the trace and retrace scan lines, $N_{\theta < \theta_{free}}$ is the number of pixels in the phase scan lines below (repulsive mode) the free phase angles, $h_{min}$ is the lowest probe position of $i^{th}$ trace and retrace height lines while $h_{global\ min}$ is the global lowest probe position computed based on all the acquired scan lines, $P(h_{trace}, h_{retrace})$ is Pearson correlation between $h_{trace}$ and $h_{retrace}$ to quantify the similarity between them. All the components inside log operator are capped minimally to the logarithmic constant $e$.

Building on these considerations, we define three rewards to quantify these characteristics based on real-time scanning data. Reward 1 (height difference) measures the agreement between the trace and retrace lines, computed as the normalized absolute difference between them. Reward 2 (distance) quantifies probe-sample proximity using the absolute height of the probe. Reward 3 (phase) detects the transition between attractive and repulsive scanning modes, calculated as the fraction of time the phase remains above the free-air phase. Reward 4 is a different way of quantifying the agreement between trace and retrace and we will use MOBO to evaluate which of the reward 1 and 4 is better at distinguishing good scans from bad ones.

Each of these rewards plays a distinct role in optimizing TM scans. The distance reward encourages the probe to remain close to the sample surface rather than oscillating in free space. The height difference reward further refines this proximity by ensuring that the probe closely follows the surface topography. Meanwhile, the phase reward prevents excessive probe-sample contact that could push the system into the repulsive mode. Consequently, the height difference reward favors a combination of high drive amplitude and low setpoint, as illustrated in Figure 1a. The distance reward assigns higher values to configurations with a low setpoint, as shown in Figure 1b. In contrast, the phase reward promotes lower drive amplitudes and higher setpoints, as depicted in Figure 1c.

As a result, the height difference and distance rewards are expected to share an overlapping optimal solution, leading to a trivial Pareto point (Figure 1e). However, the phase reward competes with both the height difference and distance rewards, necessitating trade-offs among them, as demonstrated in Figure 1f-g.

**Parameter space** – Drive amplitude, Setpoint amplitude, and I Gain.

Here we choose the drive amplitude, setpoint amplitude, and integral (I) gain as the controlling parameters to optimize. The range of drive amplitude is chosen according to the safe-seeding routine to be 0 to 100 nm as shown in Figure 2, with setpoint ranging from 0.1 % to 90 % of the free-air amplitudes. I gain in the PI loop extents from 30 to 200, chosen according to human heuristic about the SPM system used in this work.

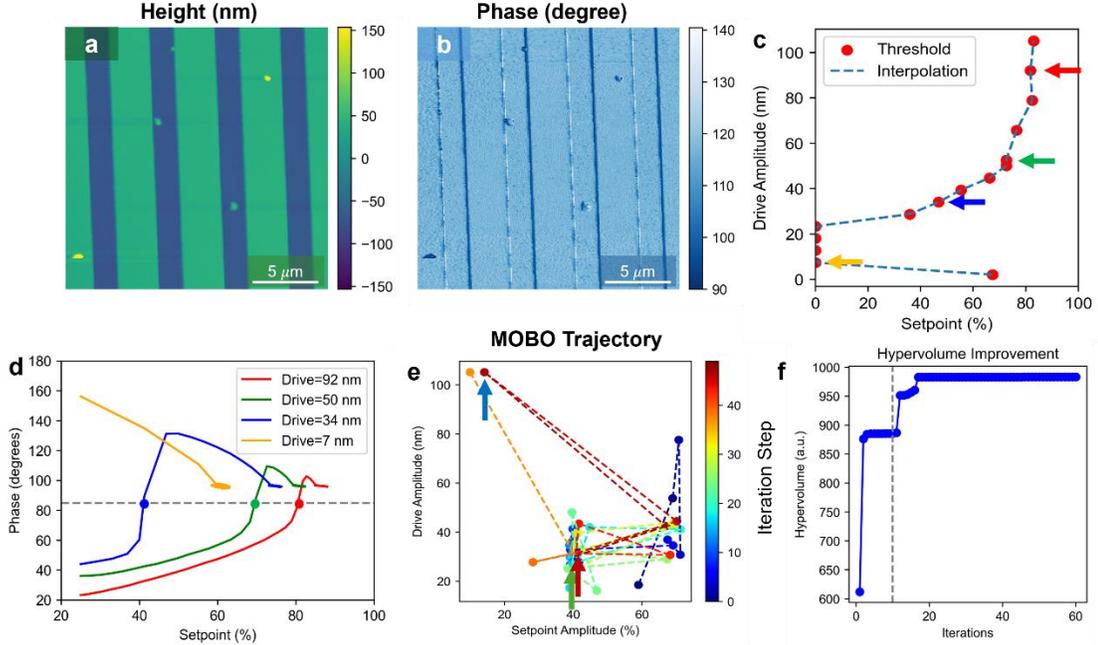

**Figure 2. Safe seeding and stopping criteria | a-b,** MOBO of tapping mode on silicon calibration grating sample with 100 nm deep trenches. Here the drive amplitude, setpoint amplitude, and I Gain are chosen to be the controlling parameters to optimize. There are 10 initial random seeding points and 50 explorative active learning steps in this experiment, and the acquisition function of q-Noisy Expected Hypervolume Improvement (qNEHI) is used to determine the next parameters to try at each step. A fresh Tap300Al-G is used to optimize scan at 1 Hz in a 20 $\mu$m area. **a-b,** the resulting **a** height and **b** phase maps taken with the optimal parameters given by the MOBO workflow. **c-d,** safe seeding with force-distance measurement. **c,** the threshold setpoints at different drive amplitudes are derived from corresponding FD curves as the intersection setpoint between phase-setpoint curve and the free-air phase in **d**. The measured threshold setpoints are interpolated in the full drive amplitude range and only setpoints greater than the threshold setpoints are considered as safe parameters. **e,** trajectory of MOBO exploration projected from 3D parameter space to the drive amplitude vs. setpoint 2D space with the I gain set to be its optimal value of 66.78. Color map represents the iteration number of steps connected by dashed lines with corresponding colors. **f,** hypervolume gain curve at different iteration steps for the MOBO workflow. The dashed gay lines separate the initial seeding steps from the active learning steps.

### III. Optimization process

The effectiveness of MOBO and its corresponding Pareto front analysis are first demonstrated on a silicon calibration grating sample from Oxford Instruments Asylum Research, as shown in Figure 2a-b. Before optimization, a safe seeding process is implemented to ensure robust exploration of the parameter space. This process begins with a series of force-distance measurements to identify the threshold setpoints at which the system transitions from the attractive to the repulsive regime. The transition is characterized by a sudden phase drop from above to below the free-air phase. Only setpoints exceeding this threshold—corresponding to the attractive mode—are included in the optimization, as illustrated in Figure 2c-d.

To initialize the MOBO model, 10 random seeding points are sampled from the safe parameter space to train the Gaussian process (GP) models. During active learning iterations, the three reward functions are computed and averaged over five neighboring real-time trace and retrace scan lines. Subsequently, three separate GPs are trained to model these reward functions as a function of the control parameters: drive amplitude, setpoint, and integral gain (I gain). From these models, the acquisition function q-Noisy Expected Hypervolume Improvement (q-NEHI) [50, 51] is computed to guide the optimization process. A reference point is selected in the hyperparameter reward space at a fixed distance below the maximum of each reward, allowing q-NEHI to quantify the potential improvement in hypervolume across the parameter space.

At each iteration, a Pareto front is computed with respect to the reference point, and the next experimental parameters are selected by maximizing q-NEHI. The optimization process continues until either the maximum number of iterations is reached or the hypervolume improvement converges. Post-analysis of the MOBO trajectory (Figure 2e) and the hypervolume gain curve (Figure 2f) suggest that optimal solutions are typically identified within approximately 10 exploration steps. Upon completion of the optimization, a full topography scan is conducted using the optimized parameters to validate the quality of the final scan.

## IV. Analysis of the Pareto front

The Pareto scatter plots in Figure 3a-c confirm that the height difference reward exhibits a trivial Pareto point with the probe-sample distance reward, as indicated by the green arrow. The competition between the phase reward and the other two rewards appears less pronounced, as evidenced by the relatively small variations in the phase reward across most of the scattered points in Figure 3g-i. This limited variation arises because only a small fraction of pixels (~3%) contribute to the phase reward, primarily at step edges, which significantly diminishes its influence in the optimization process.

Although there is a set of controlling parameters that can maximize both the reward of height difference and the reward of tip-sample distance as labeled by the green arrow in Figure 3a, it does not maximize the reward of phase. Instead, the final optimal solution pointed by the red arrow has slightly better reward of phase as shown in Figure 3b-c and verified by less phase pixels below the free-air phase in Figure 3f compared to Figure 2e. The Pareto front scatter plots further illustrate that to reach this optimal solution, indicated by the red arrow, MOBO sacrifices the tip-sample distance reward in the trade-off between the reward of height difference and the reward of tip-sample distance, as indicated by the red arrow laying in the lower right of Figure 3a. This choice agrees with human heuristics that the tip-sample distance could be affected by the tilt of the sample and thus less relevant to the overall scan quality compared to the reward of height difference.

Such conclusion is further supported by the predicted distribution of the three rewards shown in Figure 3g-i. The reward of phase in Figure 3i shows broader maxima compared to the other two rewards, agreeing with our conclusion from Pareto front analysis that the reward of phase is less selective due to the lack of step edge pixels in the calibration grating sample. Notably, the maxima of the height difference reward (Figure 3g) and the phase reward (Figure 3i) are in closer proximity within the parameter space, whereas the optimal solution for the tip-sample distance reward (Figure 3h) is located in a different region under the selected I gain parameter. This spatial separation reinforces the observed trade-offs in the MOBO process.

Finally, Figure 3d-f presents three representative trace and retrace scan lines of height and phase, with their corresponding control parameters (Figure 2e) and reward values (Figure 3g-i) highlighted by arrows in matching colors.

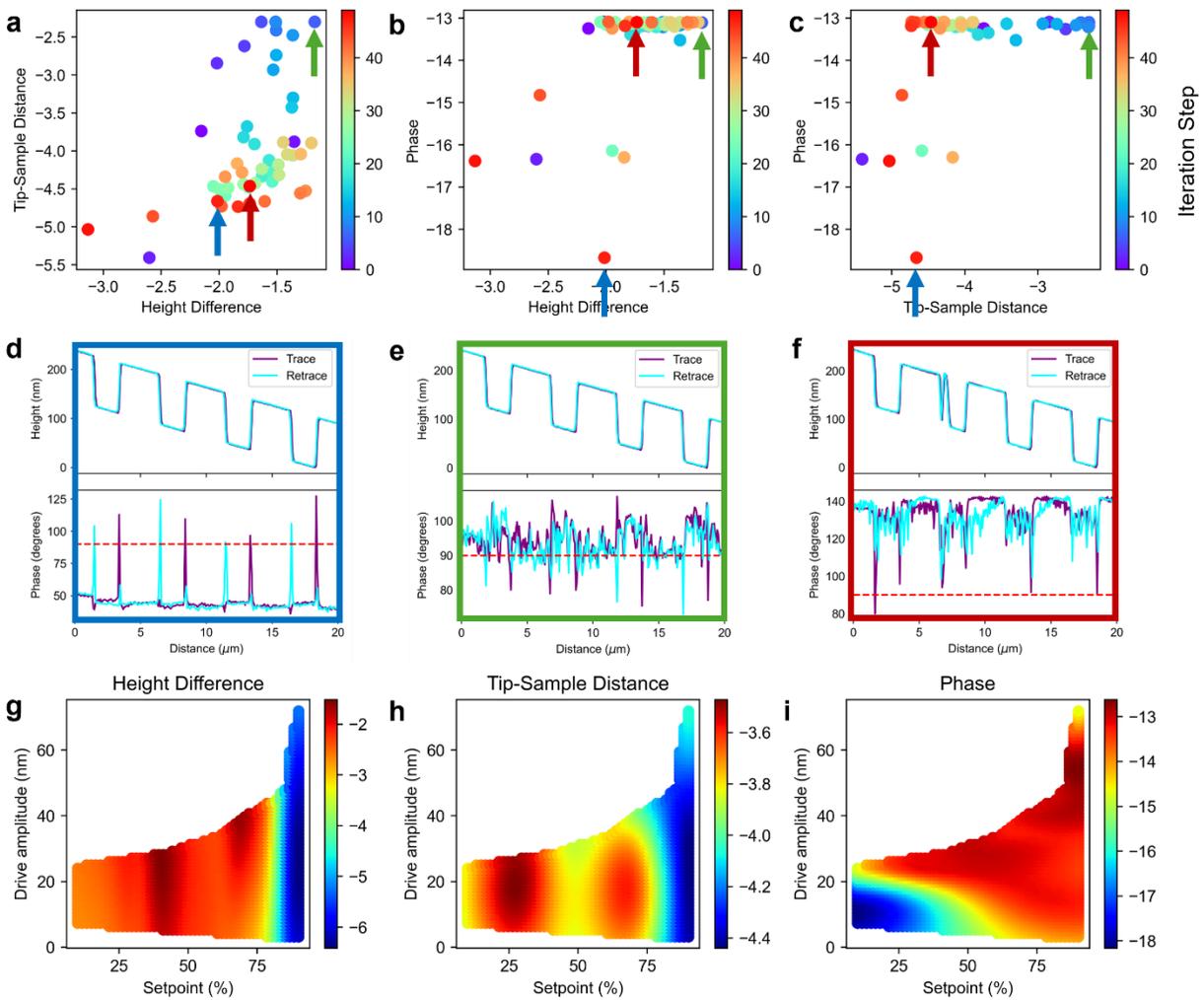

**Figure 3. Predicted rewards and Pareto front of MOBO on the calibration grating | a-c,** Pareto scatter plots for **a** height difference vs. tip-sample distance, **b** height difference vs. phase,

and **c** tip-sample distance vs. phase. The color scale represents the iteration steps in the MOBO. The height difference shows collaborative behavior with the reward of tip-sample distance. In contrast, the reward of phase shows competing behavior with the reward of height difference and reward of tip-sample distance. **d-f,** characteristic scan trace and retrace lines of height and phase for **d** pressing too hard, **e** pressing slightly too hard, and **f** the optimal solution. The colored outline matches the colored arrows in a-c to show their corresponding location on the Pareto scatter plot. **g-i,** the predicted distribution of **g** height difference, **h** tip-sample distance, and **i** phase rewards in the drive amplitude vs. setpoint space at the 50$^{th}$ step in the MOBO of the calibration grating sample in Figure 2. Detailed controlling parameters used in this figure are summarized in Table S1.

## V. Pareto front tests the definition of the reward

In Figure 4, instead of employing the three rewards introduced in Figures 1 and 2 within the MOBO process, a fourth reward is introduced to demonstrate how the Pareto front can be leveraged to validate the definition of rewards. In addition to absolute height difference, the Pearson correlation between trace and retrace scan lines—referred to as the similarity reward—can be used to quantify their alignment.

The resulting Pareto scatter plot between the height difference reward and the similarity reward in Figure 4c highlights that height difference is a more effective metric for quantifying the alignment of trace and retrace scan lines. This conclusion is based on two key observations. First, the height difference reward exhibits a larger span compared to the similarity reward, allowing it to contribute more significantly to hypervolume gain and thus carry greater weight in trade-offs with other rewards. In contrast, the similarity reward's narrow range of variation makes it less influential in the acquisition function, leading to its underrepresentation in the optimization process. Second, the similarity reward is clustered predominantly around a value of –1, despite the MOBO process exploring a broad range of scan qualities. In contrast, the height difference reward is smoothly distributed across its full range, further demonstrating its superior ability to capture variations in trace-retrace alignment.

A well-defined reward function for scan quality should exhibit both a broad dynamic range, ensuring its contribution to hypervolume improvement, and a well-separated, continuous distribution, allowing for effective differentiation of scan qualities. The Pareto scatter plot thus serves as a valuable tool for evaluating the effectiveness of different reward definitions.

The Pareto scatter plots of the remaining three rewards (Figure 4d-f) further illustrate trade-offs in the water droplet system. Notably, the height difference reward shares similar optimal parameters with the phase reward, as evidenced by their collective behavior in Figure 4e—where height difference can be maximized with minimal compromise in phase reward. This relationship is further corroborated by Figure S2c and S2e. Consequently, the tip-sample distance reward is sacrificed in the optimization trade-off to maximize overall hypervolume improvement, as indicated in Figure 3d.

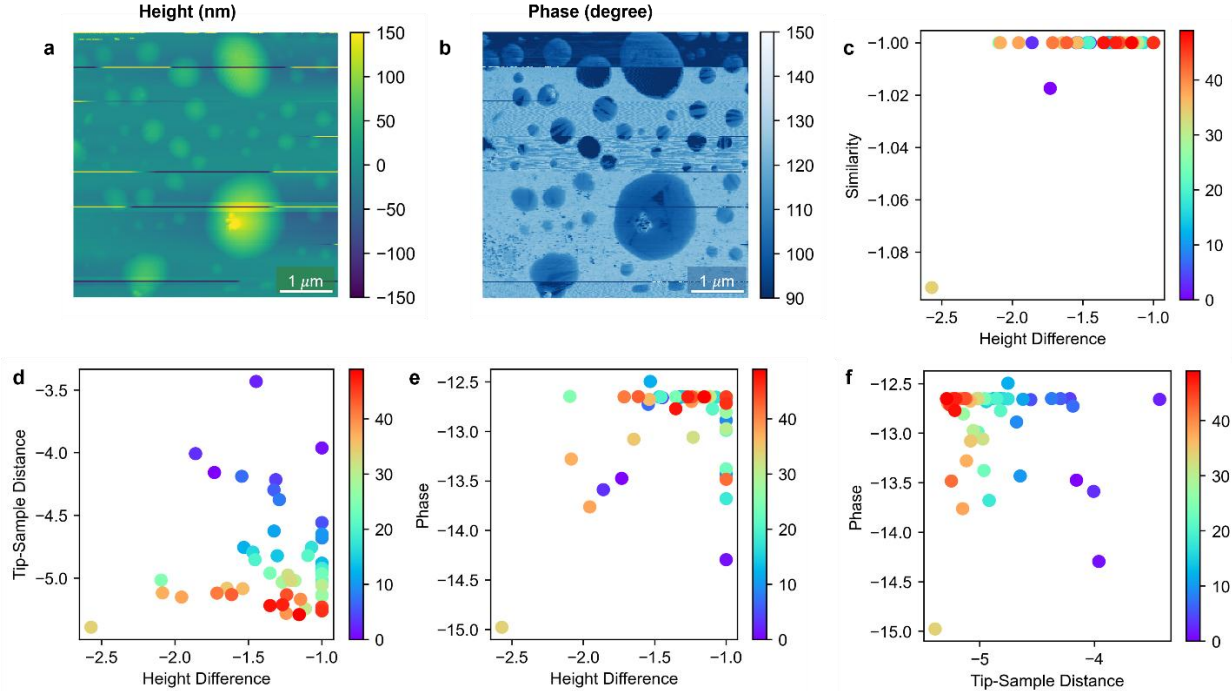

**Figure 4. MOBO on more challenging sample and validation of reward designs** | MOBO of tapping mode on water droplet sample on mica substrate. Here the drive amplitude, setpoint amplitude, and I Gain are chosen to be the controlling parameters. There are 10 initial random seeding points and 50 explorative active learning steps in this experiment, and the acquisition function of qNEHI is used. A fresh Tap300Al-G is used to optimize scan at 1 Hz in a 5 $\mu$m area. **a-b,** the training maps of **a** height and **b** phase are taken with the different parameters during the MOBO process. These maps correspond to the scan lines taken with 1-8[th] iterations in the MOBO trajectory plot in Figure S2a. **c,** the Pareto scatter plot between the reward of height difference and the reward of similarity, which are defined to quantify the same consideration of how well the trace is aligned with the retrace scan line. The reward of height difference is better because it shows more significant reward improvement compared to the similarity, and it shows smoother change in the whole MOBO process which guarantees a larger distinction between good and bad scans. **d-f,** Pareto scatter plot for the rewards of height difference, tip-sample distance, and phase. Compared to similar plots in Figure 3, the reward of phase shows better resolution.

### VI.    MOBO as a Human-in-the-Loop Optimization Framework

MOBO provides a mechanism for incorporating human decision-making into the optimization process, particularly in balancing trade-offs between competing rewards along the Pareto front. This capability is crucial in scientific experiments where human expertise is needed to prioritize specific experimental outcomes. This is particularly useful in the situation where some rewards are more important than others, or where it requires human intervention as there is no satisfactory optimal solutions under given circumstances.

For instance, in Figure 5, we first perform a MOBO-driven optimization of tapping mode on a water droplet sample, yielding three GP models that predict the distribution of the three rewards across the 3D parameter space. Typically, the acquisition function is computed based on the predictions and uncertainties of these GP models to determine the optimal control parameters. However, MOBO enables dynamic reweighting of rewards, allowing the acquisition function to favor particular rewards by adjusting their relative importance.

Figure 5c-d illustrates how modifying the weight of the phase reward and height difference reward influences the optimal solution in the 2D parameter space. The corresponding scan lines (Figure 5e-i) demonstrate that decreasing the weight of the phase reward has minimal impact on the height and phase scan lines. This is consistent with the observed clustering of solutions in the lower left of Figure 5c, suggesting that a steep gradient in the parameter space leads to substantial hypervolume changes with small parameter shifts. This effect arises from the saturation of the phase reward: once all phase pixels remain above the free-air phase, further improvements become marginal. In contrast, increasing the weight of the phase reward (Figure 5g) effectively lowers the phase at the expense of reduced alignment between the height trace and retrace.

Altering the weight of the height difference reward has a more pronounced effect on scan quality. As shown in Figure 5h, decreasing its weight lowers the phase value by increasing the probe-sample distance, but this comes at the cost of significantly worse alignment between trace and retrace. Conversely, increasing the height difference reward weight (Figure 5i) pushes the probe closer to the sample surface, resulting in only marginal improvements in trace-retrace alignment while negatively impacting the phase reward—evidenced by phase pixels transitioning from the attractive to the repulsive mode.

Thus, MOBO allows for quantitative and controlled adjustments to optimization objectives, aligning outcomes with human preferences. More importantly, these modifications are made with explicit quantification of their impact on scan quality, providing an objective measure of trade-offs between rewards. Traditionally, human operators adjust control parameters based on intuition and empirical experience to achieve desired scan properties. With MOBO, instead of manually tuning individual parameters, operators can now directly specify which aspect of the scan to improve by adjusting the corresponding reward weights in a systematic and data-driven manner.

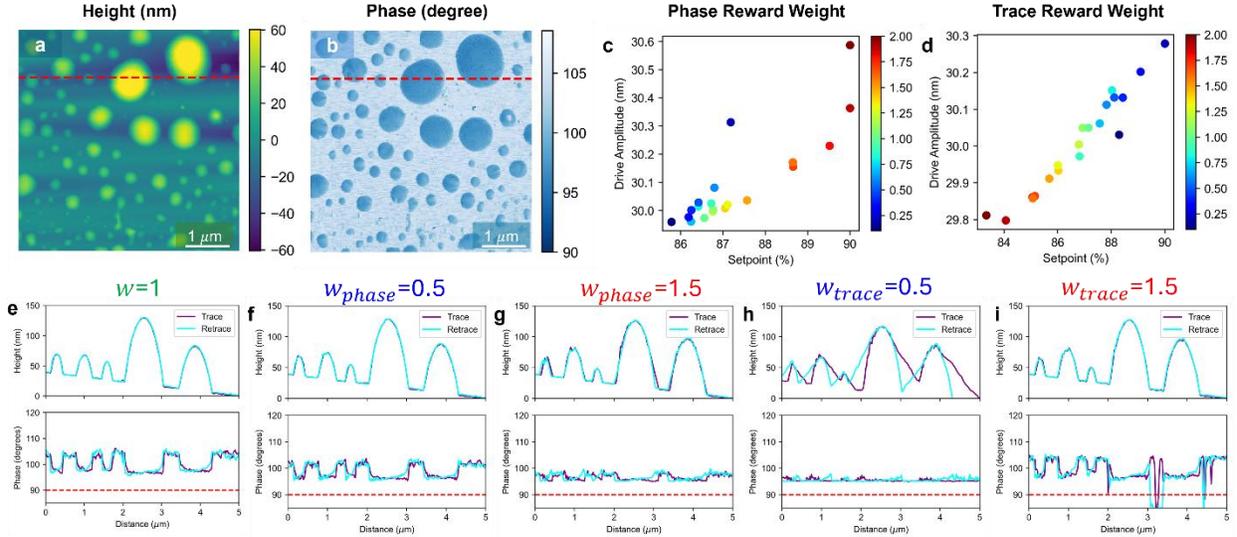

**Figure 5. Bring humans to decide the trade-offs on the Pareto front |** MOBO of tapping mode on water droplet sample on mica substrate. Here the drive amplitude, setpoint amplitude, and I Gain are chosen to be the controlling parameters. There are 20 initial random seeding points and 50 explorative active learning steps in this experiment, and the acquisition function of qNEHI is used. A Tap300Al-G is used to optimize scan at 1 Hz in a 5 $\mu$m area. **a-b,** the resulting **a** height and **b** phase maps taken with the optimal parameters given by the MOBO process. **c,** after 20 seeding plus 50 explorative steps, different weights (0.1 to 2.0 with 0.1 increment) are multiplied to the reward of phase before retraining the MOBO GP models. The new optimal controlling parameter of drive amplitude, setpoint, and I gain are given by the qNEHI acquisition function at each added weight. Here the updated optimal solutions are plotted in the drive amplitude vs. setpoint space, showing how the optimal solution evolves with different emphasis on the phase reward. **d,** a similar plot for added weight for the reward of trace. After the extra weights are added, topography maps with new optimal controlling parameters are taken: **e-i,** the trace and retrace scan lines of height and phase taken along the red dashed lines in a and b with optimal controlling parameters predicted by **e** default weights on all the rewards, **f** weight of 0.5 for the reward of phase, **g** weight of 1.5 for the reward of phase, **h** weight of 0.5 for the reward of trace, and **i** weight of 1.5 for the trace. MOBO offers a way to optimize controlling parameters with control on how much each reward will be improved quantitatively. Detailed controlling parameters used in this figure are summarized in Table S1.

### VII. Conclusion

We demonstrate that Multi-Objective Bayesian Optimization effectively balances multiple uncertain rewards, enabling FAIR (Findable, Accessible, Interoperable, and Reproducible) access to high-quality and reproducible experimental results. Beyond optimizing system performance,

MOBO provides valuable insights into the interplay between competing rewards and incorporates human decision-making into the optimization process by allowing trade-offs within the Pareto front to be tailored for different scenarios.

We illustrate these principles through the automated optimization of tapping mode in Scanning Probe Microscopy (SPM), where three reward functions—derived either from the underlying physics of the system or from human heuristics—guide the optimization. Our results show that MOBO rapidly converges to optimal control parameters, producing high-quality, reproducible scans while quantifying the relationships between competing rewards. The Pareto front analysis not only helps to refine the design and validation of reward functions but also enables human operators to actively participate in the decision-making process.

By assigning weights to different rewards, MOBO allows human operators to precisely control the trade-offs between objectives, transforming each reward into a quantifiable tuning parameter. This framework shifts experimental optimization from intuitive, trial-and-error approaches to a systematic, data-driven strategy, enhancing both efficiency and reproducibility in complex scientific experiments.

**MOBO implementation and instrument control**

For all the MOBO workflows in this work, we limit the parameter space resolution to 100 x 100 x 50 so that the workflow is lightweight enough to run on a local computer with central processing unit (CPU) only. The surrogate Gaussian Process (GP) model was incorporated using gpytorch [52]. MOBO is implemented in BOTorch [53].

The SPM control is achieved by our home built open-source Python interface library, AESPM [54]. This library not only enables real-time operating the SPM system local or remotely with code the same way as human operators but also has access to the intermediate data like trace and retrace scan lines in all the channels which are essential for fast optimization presented in this work.

AESPM is an open-source SPM-Python interface library. It can be found in the following link with detailed examples and tutorial notebooks: https://github.com/RichardLiuCoding/aespm

SpmSimu is an open-source SPM scanning simulator. It can be found in the following link with detailed examples and tutorial notebooks: https://github.com/RichardLiuCoding/spmsimu

To help readers understand and reproduce the results in this work, we have provided an open-source Python notebook for the tutorial presented in the Appendix 1: https://github.com/RichardLiuCoding/Publications/blob/main/MOBO%20Tutorial_v2.ipynb

We have also prepared a full MOBO workflow based on our SpmSimu simulator that readers can modify to build automated workflow on their own instruments:

https://github.com/RichardLiuCoding/Publications/blob/main/AC%20MOBO%20based%20on%20SPM%20simulator_v5.ipynb

**Acknowledgement**

This work was supported by the Center for Advanced Materials and Manufacturing (CAMM), the NSF MRSEC center.

We thank Astita Dubey and Mahshid Ahmadi for providing us with CBI microcrystals for challenging our MOBO workflow, and Roger Proksch for discussing safe seeding.

## I. Introduction to Multi-objective Bayesian Optimization

The Pareto front is the set of non-dominated solutions in the objective space, meaning no other solution is strictly better in *all* objectives at the same time. For example, in the two-objective maximizing problem shown in Figure S1, a point in the hyperparameter (reward) space is a Pareto point, also called non-dominated point, if and only if there are no other points on its right horizontally **and** there are no other points on its top vertically. When the two rewards have overlapping optimal solutions, their Pareto front will collapse into a trivial Pareto point, as shown in Figure S1d. Once the two rewards have different optimal parameters, there will be trade-offs involved on the nontrivial Pareto front as shown in Figure S1e-f.

$$\text{Reward 1} = \exp(-(x_1 - 0.35)^2 - (x_2 - 0.65)^2) \\ - \exp(-(x_1 - 0.65)^2 - (x_2 - 0.35)^2) \tag{5}$$

$$\text{Reward 2} = \exp(-(x_1 - 0.35)^2 - (x_2 - 0.35)^2) \\ - \exp(-(x_1 - 0.65)^2 - (x_2 - 0.65)^2) \tag{6}$$

$$\text{Reward 3} = -\text{Reward 1} \tag{7}$$

Here $x_1$ and $x_2$ both range from 0 to 1.

In the MOBO process, the measured Pareto front is computed based on all the measured reward values. In the meantime, multiple individual GPs are used to predict the distribution and uncertainty of each reward in the whole experiment parameter space, based on which the acquisition function can be calculated. In the example of qNEHI, the expected hypervolume gain is computed in the hyperparameter space, considering the expected improvement in each reward and their corresponding uncertainty in the parameter space. Therefore, as shown in Figure S1d-f, the next set of parameters to try is determined by maximizing such expected hypervolume improvements (rewards gain), or equivalently how far the new measurement can push Pareto front away from the reference point to the "upper right" according to GP predictions.

In Figure S1, we show the distribution of Pareto front in the hyperparameter (reward) space in panel e, and parameter space in panel g for reward 1 vs. reward 2 MOBO problem. Notably, the Pareto front appears as an arc connecting the maximums of reward 1 and 2 along their overlapping gradient direction.

We subsequently examine the effect of the two different ways of biasing the MOBO choice in the acquisition function. Here we take 50 random seeding points in the $x_1 - x_2$ parameter space and then modify the acquisition function to give different predictions on the optimal solutions. In Figure S1h, we show the consequence of shifting the reference point. When the reference point is along the diagonal line (ref 1 in Figure S1e), the resulting optimal solution locates in the middle of the maximums of reward 1 and 2 (blue point in Figure S1h), showing no preference on neither reward 1 nor reward 2. When we lower the reference point for reward 1 (the orange point in Figure S1e), the new optimal solution given by MOBO shifts away from the maximum of this reward along the Pareto front, indicated by the orange point in Figure S1h. Therefore, to favor a reward, one needs to increase the corresponding reference point closer to the maximum of that reward.

The other method is to multiply weights directly to the measured rewards before feeding them to the acquisition function. For example, in Figure S1i, we show the results of multiplying the reward 1 with weights ranging from 0.75 to 1.25 on the optimal solutions predicted by the same MOBO. When a weight of greater than 1 is multiplied to a reward, the optimal solution will be moved away from its maximum along the Pareto front. In other words, adding a larger weight to a reward will suppress it in the MOBO.

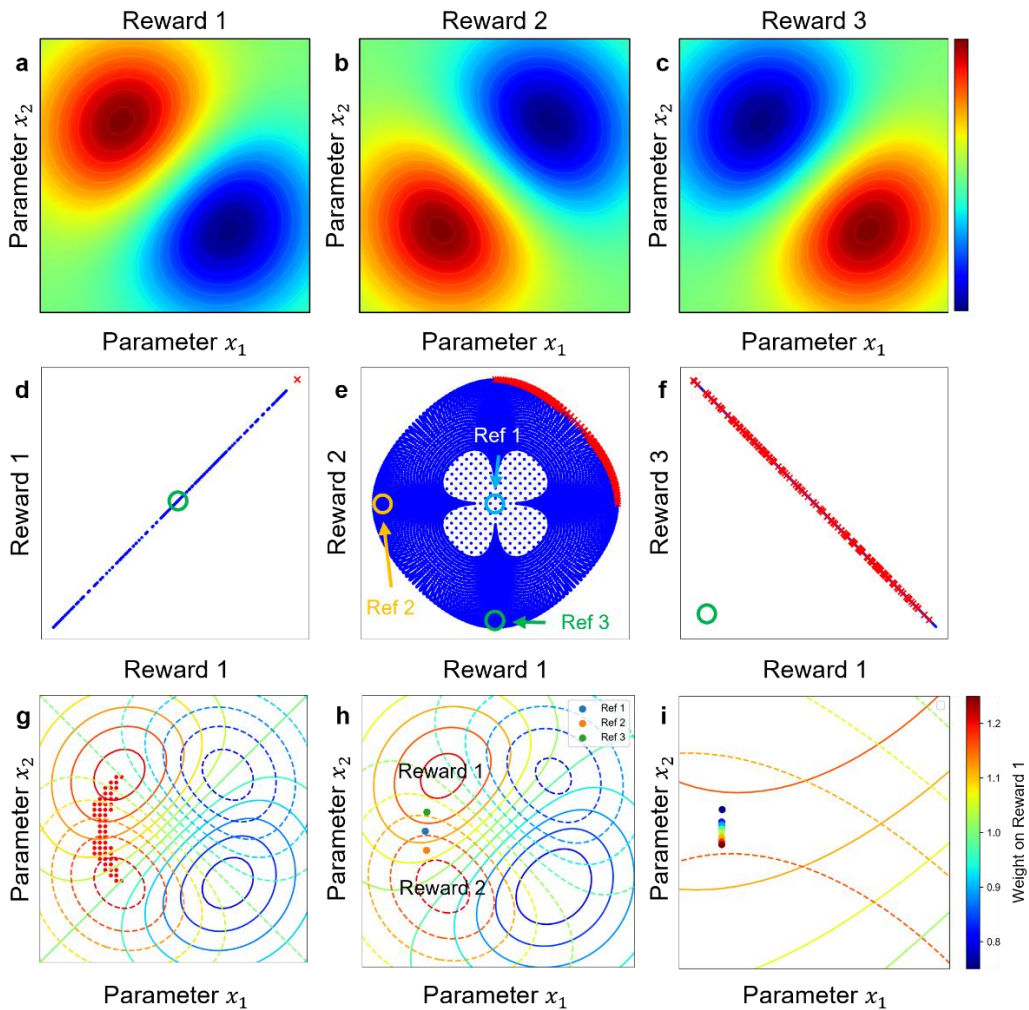

**Figure S1. Overview of MOBO and Pareto front | a,** an example distribution of reward in the parameter space. **b,** reward 2 is 90-degree rotated counter-clockwise with respect to reward 1 so

that it's orthogonal to the reward 1. **c,** reward 3 is 180-degree rotated to reward 1 and it's opposite to the reward 1. **d,** Pareto scatter plot for reward 1 and reward 1 itself. When the two rewards have optimal solution overlapping in the parameter space, the Pareto front collapses into a trivial Pareto point (the red cross mark) that optimizes both rewards at the same parameter. **e,** In the case that the optimal solution of reward 1 is different from that of reward 2, there will be trade-offs between reward 1 and 2 on the Pareto front. **f,** When the two rewards are opposite to each other, all the points in the Pareto scatter plot will be Pareto front points, meaning that at each point, there is no another point that can further improve any of the rewards without sacrificing the other reward. The green circles in d-f show an example choice of reference in the hyperparameter (rewards) space. **g,** the distribution of Pareto front in the parameter space. Here the reward 1 and reward 2 are plotted as solid and dashed contours together, with the red dots representing the same Pareto front in the hyperparameter space of panel e. **h,** the effect of shifting the reference point on the MOBO choice. Lowering the reference point for a reward will move the optimal solution away from it along the Pareto front. **i,** the effect of adding weight to the rewards in the acquisition function. Adding extra weight (greater than 1) to a reward will also move the optimal solution away from it along the Pareto front.

| Figures | Drive Amplitude (nm) | Setpoint (%) | I Gain (a.u.) |
|---|---|---|---|
| **Figure 2-a,b,i** | 30.11 | 40.32 | 66.78 |
| **Figure 2-g** | 105.18 | 14.18 | 30.00 |
| **Figure 2-h** | 36.10 | 39.58 | 73.92 |
| **Figure 4-a,b,e** | 29.90 | 84.99 | 67.56 |
| **Figure 4-f** | 30.01 | 86.43 | 66.64 |
| **Figure 4-g** | 30.06 | 87.91 | 69.43 |
| **Figure 4-h** | 30.08 | 87.79 | 69.70 |
| **Figure 4-i** | 29.84 | 84.94 | 62.73 |

**Table S1. Detailed controlling parameters for each plot.**

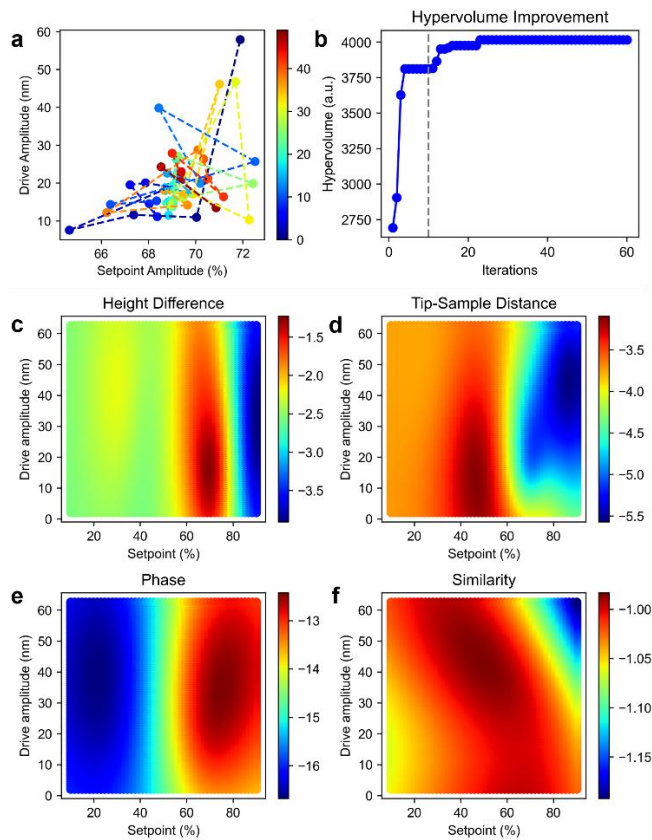

**Figure S2. Supplementary plots for the MOBO of droplet sample in Figure 3 | a,** trajectory of MOBO exploration projected from 3D parameter space to the setpoint vs. drive amplitude 2D space with the I gain set to be its optimal value. **b**, hypervolume gain curve at different iteration steps for the MOBO training workflow. The dashed gay lines separate the initial seeding steps from the active learning steps. **c-f,** the predicted distribution of **c** height difference, **d** tip-sample distance, **e** phase, and **f** similarity rewards in the setpoint – drive amplitude space at $50^{th}$ in the MOBO.

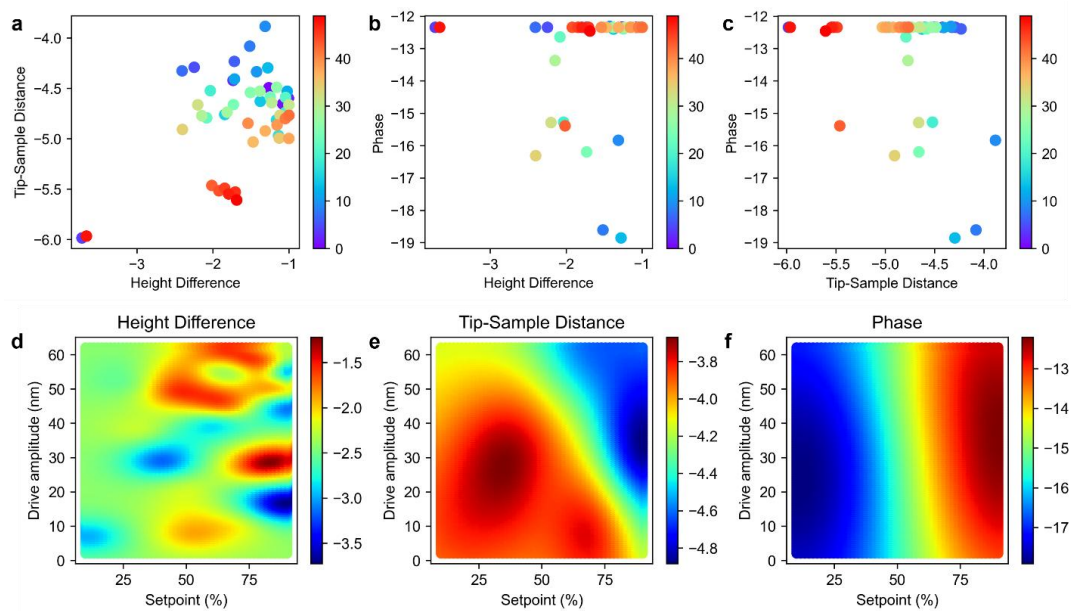

**Figure S3. Full Pareto scatter plot for the MOBO of droplet sample in Figure 4 and predicted distribution of the three rewards at the final step of MOBO.**

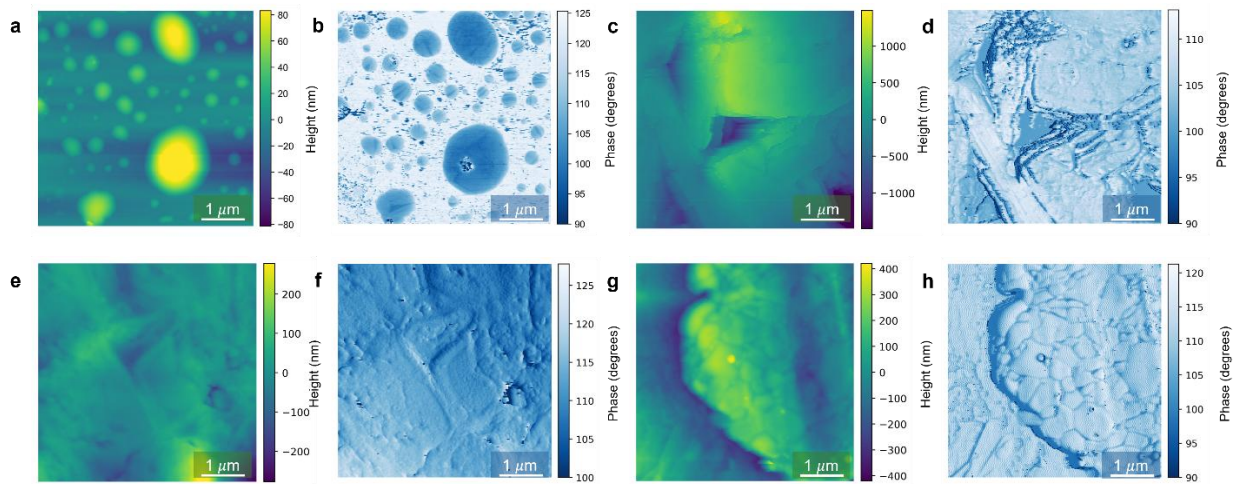

**Figure S4. Testing of MOBO on more samples | a-b,** height and phase maps taken with optimal parameters given by MOBO for water droplet sample. Similar plots are shown for **c-d** $Cs_3Bi_2I_9$ (CBI) microcrystals with a deep hole, **e-f** flat surface on CBI microcrystals, and **g-h** a grain boundary on CBI microcrystals. There are 10 initial random seeding points and 50 explorative active learning steps in this experiment, and the acquisition function of q-Noisy Expected Hypervolume Improvement (qNEHI) is used to determine the next parameters to try at each step. A fresh Tap300Al-G is used to optimize scan at 1 Hz.